\pdfoutput=1

\documentclass[journal]{IEEEtran}

\usepackage[noadjust]{cite}
\usepackage[pdftex]{graphicx}
\usepackage{xcolor}
\usepackage{pifont}
\usepackage{array}
\usepackage{amssymb}
\usepackage{url}
\usepackage{enumerate}
\usepackage{tabularx}
\usepackage{tabularray}
\usepackage[multiple]{footmisc}

\begin{document}

\title{in-Car Biometrics (iCarB) Datasets for Driver Recognition: Face, Fingerprint, and Voice}

\author{Vedrana~Krivoku\'ca~Hahn$^1$,
	    J\'er\'emy Maceiras$^2$,
	    Alain Komaty,
	    Philip Abbet,
        S\'ebastien~Marcel \\
        {\tt\small $^1$vkrivokuca@idiap.ch}, {\tt\small $^2$jmaceiras@idiap.ch} \\
        \textit{Idiap Research Institute (Martigny, Switzerland)}  
}

\maketitle

\begin{abstract}

We present three biometric datasets (\textit{iCarB-Face}, \textit{iCarB-Fingerprint}, \textit{iCarB-Voice}) containing face videos, fingerprint images, and voice samples, collected inside a car from 200 consenting volunteers. The data was acquired using a near-infrared camera, two fingerprint scanners, and two microphones, while the volunteers were seated in the driver's seat of the car. The data collection took place while the car was parked both indoors and outdoors, and different ``noises'' were added to simulate non-ideal biometric data capture that may be encountered in real-life driver recognition. Although the datasets are specifically tailored to in-vehicle biometric recognition, their utility is not limited to the automotive environment. The \textit{iCarB} datasets, which are available to the research community, can be used to: (i) evaluate and benchmark face, fingerprint, and voice recognition systems (we provide several evaluation protocols); (ii) create multimodal pseudo-identities, to train/test multimodal fusion algorithms; (iii) create Presentation Attacks from the biometric data, to evaluate Presentation Attack Detection algorithms; (iv) investigate demographic and environmental biases in biometric systems, using the provided metadata. To the best of our knowledge, ours are the largest and most diverse publicly available in-vehicle biometric datasets. Most other datasets contain only one biometric modality (usually face), while our datasets consist of three modalities, all acquired in the same automotive environment. Moreover, \textit{iCarB-Fingerprint} seems to be the first publicly available in-vehicle fingerprint dataset. Finally, the \textit{iCarB} datasets boast a rare level of demographic diversity among the 200 data subjects, including a 50/50 gender split, skin colours across the whole Fitzpatrick-scale spectrum, and a wide age range (18-60+). So, these datasets will be valuable for advancing biometrics research. 

\end{abstract}


\section{Introduction}

The collection of the three biometric datasets that are presented in this paper (\textit{iCarB-Face}, \textit{iCarB-Fingerprint}, and \textit{iCarB-Voice}) was motivated by a requirement to investigate in-vehicle driver identification using face, fingerprint, and voice biometrics. The goal was to use the datasets to evaluate and benchmark different biometric systems, to gain insight into which biometric modality (and system) would be most suitable for in-vehicle driver identification. An important aim of the data collection, therefore, was to simulate conditions that would make it difficult to capture ``clean'' biometric data in practice (e.g., noises, uncontrolled lighting, finger dryness), thus challenging the recognition capabilities of the investigated biometric systems. Furthermore, to ensure a fair evaluation across different demographics, it was necessary for the biometric data to be gender balanced, and to represent diverse skin colours and ages. Existing biometric datasets were found to be inadequate for these requirements, due to several reasons, such as: small number of subjects and lack of demographic diversity, no datasets of all three biometric modalities acquired in the same (automotive) environment, and data not having been captured in all the conditions of interest.  

The biometric datasets presented in this paper (\textit{iCarB-Face}, \textit{iCarB-Fingerprint}, and \textit{iCarB-Voice}) thus represent a valuable contribution to the research community, for a number of important reasons:

\begin{itemize}

\item \textbf{Ethical, legal, and diverse acquisition:} The combination of factors considered in acquiring the \textit{iCarB} datasets distinguishes them from other publicly available biometric datasets.  Firstly, our datasets are ethically and legally clean, having been acquired from consenting human volunteers. This sets them apart from widely used web-scraped biometric datasets, collected without the data subjects' consent. Secondly, our data was obtained from real people, as opposed to being synthetically generated, so our datasets are more accurate representations of real biometric data distributions. Thirdly, our datasets were collected from a gender-balanced cohort of 200 people, with skin colours across the whole Fitzpatrick-scale spectrum, and spanning a wide age range (18 to 60+). This level of demographic diversity, combined with the substantial number of real, consenting data subjects, is extremely rare in public biometric datasets. Finally, our data was captured in various scenarios to simulate different variabilities that may be encountered in real-life biometric recognition. So, the datasets are useful for in-depth (and consensual) evaluations of face, fingerprint, and voice biometric systems. 

\item \textbf{Benchmarking potential:} The \textit{iCarB} datasets can be used to evaluate and benchmark different face, fingerprint, and voice recognition systems. The fact that the data was collected in various ``noisy'' scenarios will be useful for understanding challenges that are likely to affect the accuracy of in-vehicle biometric recognition systems. This will help towards developing more accurate and resilient biometric systems for driver recognition. We provide several protocols to facilitate these evaluations. The datasets can also be combined to create multimodal pseudo-identities, which could be used to train and/or test multimodal fusion algorithms (e.g., score-level, rank-level, decision-level). The datasets could additionally be employed to create Presentation Attacks from the face, fingerprint, and voice data, which could be used to evaluate Presentation Attack Detection algorithms. Finally, as the datasets are accompanied by various demographic and environmental metadata (depending on the dataset), they allow for the evaluation of demographic and/or environmental bias in the corresponding biometric recognition systems. Note that, while the data are specifically tailored to in-vehicle biometric recognition, the utility of the datasets is not limited to the automotive environment.  

\item \textbf{Multimodality:} As far as we are aware, the set of \textit{iCarB} datasets represents the first effort to construct a publicly available multimodal biometric (face, fingerprint, and voice) dataset acquired inside a vehicle. It is important to emphasize that, for privacy reasons, the identities of the human data subjects do not correspond across the three datasets (e.g., ID 1 does not represent the same person in \textit{iCarB-Face}, \textit{iCarB-Fingerprint}, and \textit{iCarB-Voice}). Nevertheless, the three datasets can be used to investigate both unimodal and multimodal biometric algorithms, to gain insights into the accuracy of biometric recognition systems when they are applied to biometric data acquired in an automotive environment.   

\item \textbf{Public availability:} The \textit{iCarB} datasets are, to the best of our knowledge, the largest and most diverse publicly available in-car biometric datasets. Most other datasets contain only one biometric modality\footnote{AVICAR \cite{avicar} contains face and voice data, but the dataset does not seem to be available anymore (the link to the dataset, from the paper, is broken).}, usually the face (e.g., VFPAD \cite{vfpad}, 3DMAD \cite{3dmad}), whereas the \textit{iCarB-Face/Fingerprint/Voice} datasets consist of three biometric modalities, all acquired in the same automotive environment. Moreover, as far as we know, \textit{iCarB-Fingerprint} is the first in-vehicle fingerprint dataset to be released to the research community. The number of data subjects in existing in-vehicle biometric datasets also tends to be much smaller than in \textit{iCarB}, e.g., VFPAD\footnote{\url{https://zenodo.org/records/5839959}}: 40; 3DMAD\footnote{\url{https://sites.google.com/site/benkhalifaanouar1/6-datasets}}: 50 (daytime), 19 (nighttime); iCarB\footnote{\url{https://www.idiap.ch/en/scientific-research/data/icarb-face}}\footnote{\url{https://www.idiap.ch/en/scientific-research/data/icarb-fingerprint}}\footnote{\url{https://www.idiap.ch/en/scientific-research/data/icarb-voice}}: 200. Further, the \textit{iCarB} datasets were acquired from a wide demographic, including an equal number of males and females, which is not the case for most biometric datasets, e.g., VFPAD: 24 males + 16 females; 3DMAD: 38 males + 12 females (daytime) or 11 males + 8 females (nighttime). 

\end{itemize}

\section{Data Collection Methodology}

The biometric data contained in the \textit{iCarB-Face}, \textit{iCarB-Fingerprint}, and \textit{iCarB-Voice} datasets, consists (respectively) of face videos, fingerprint images, and voice audio samples, collected from 200 human volunteers (i.e., data subjects). The biometric sensors (i.e., camera, fingerprint scanners, and microphones) used to collect this data, were set up inside a car.  During data collection, each data subject sat in the driver's seat of the car, while the operator (i.e., data collector) sat in the front passenger seat and used an in-house laptop-based application to capture the subject's biometric data via the appropriate sensor(s). The following sub-sections describe the biometric sensor set-up and the data collection protocols. 

\subsection{Biometric sensor set-up}

The test vehicle was outfitted with various sensors to enable the acquisition of biometric data from a ``driver''. This includes: (a) a near-infrared camera with two illumination circuits; (b) two microphones; and (c) two fingerprint scanners. Fig. 1 illustrates the sensor set-up, and the sub-sections that follow provide details about the different sensors. 

\begin{figure}[!h]
	\centering
	\includegraphics[width=\columnwidth]{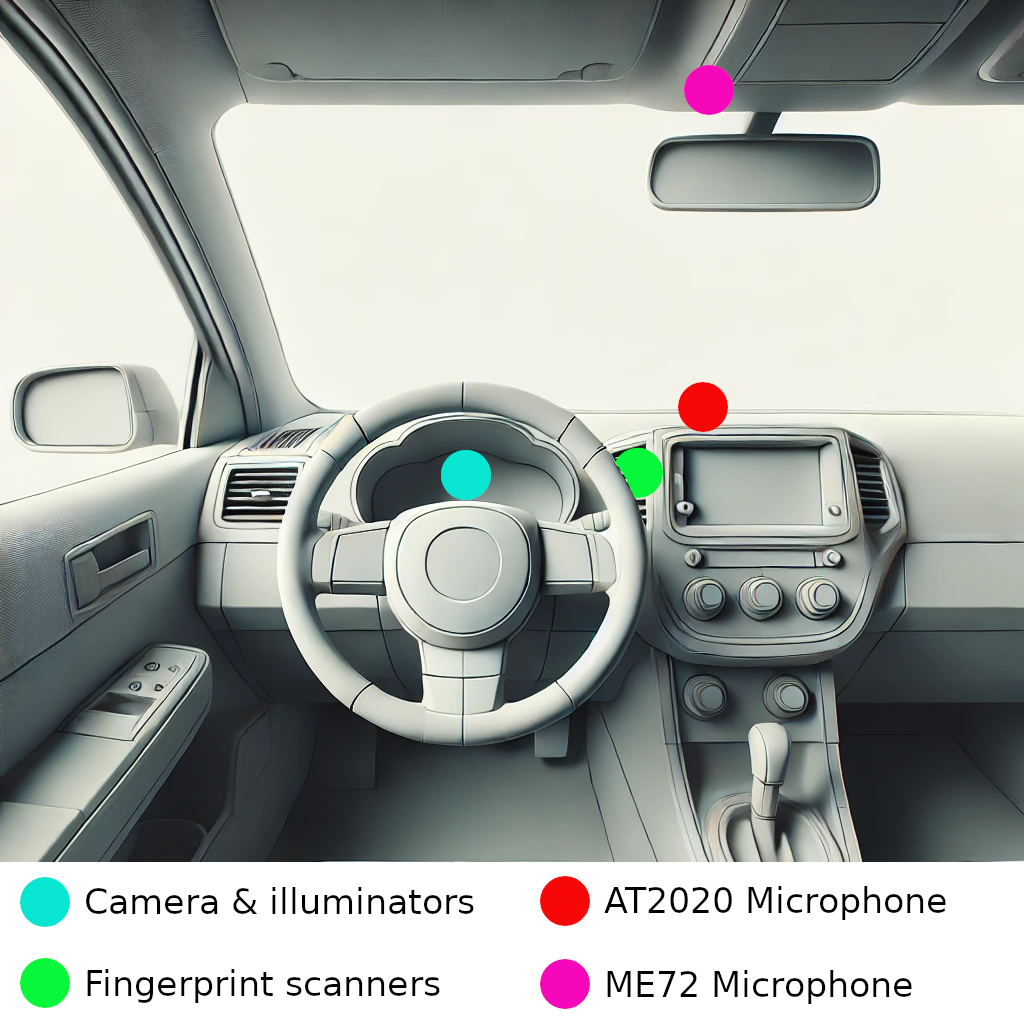}
	\caption{Biometric sensor set-up inside the car. The car cockpit was generated by Artificial Intelligence, as an NDA with our industrial partner prevents us from showing images of the sensor set-up inside the real test vehicle.}
\end{figure}

\textbf{Camera and illumination:} The Monochrome Camera MSC2-M42-1-A from Spectral Devices Inc. was used to capture videos of the data subjects' faces. In controlled lighting conditions, a custom-made near-infrared (NIR) illuminator was employed. This illuminator, positioned adjacent to the camera, features 8 LEDs with a peak wavelength of 940 nm. Fig. 2 shows an image of this camera with the illuminator, and Table I summarizes the camera’s technical specifications. 

\begin{figure}[!h]
	\centering
	\includegraphics[width=0.9\columnwidth]{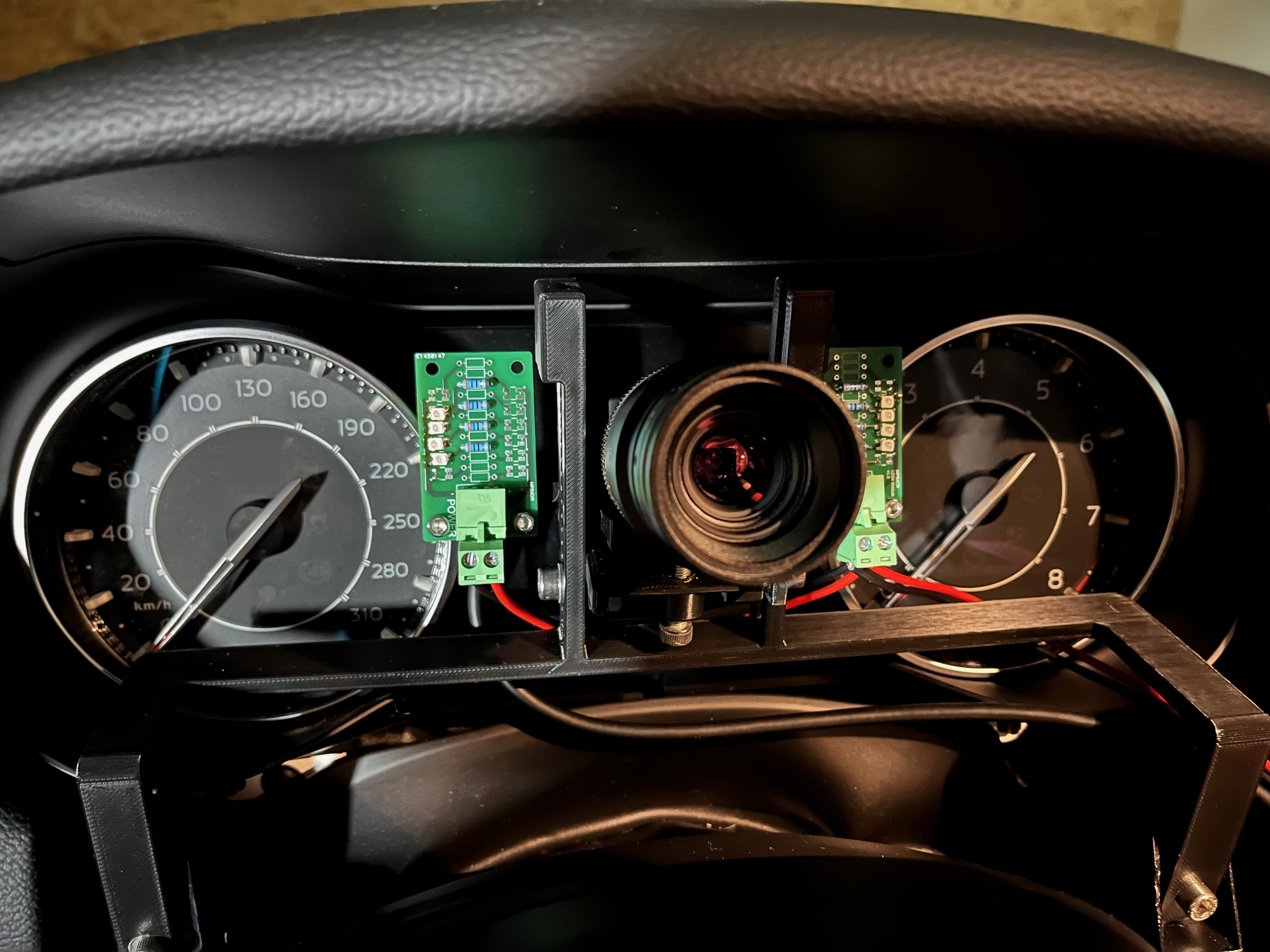}
	\caption{MSC2-M42-1-A camera and the custom NIR illuminator circuits.}
\end{figure}

\begin{table}[!h]
	\centering
	\caption{Camera specifications.}
	\begin{tabular}{|l|c|llll}
	\cline{1-2}
	\textbf{Manufacturer}               & Spectral Devices Inc. \\ \cline{1-2}
	\textbf{Model}                      & MSC2-M42-1-A          \\ \cline{1-2}
	\textbf{Dimensions}                 & 28 x 28 x 47 mm       \\ \cline{1-2}
	\textbf{Sensor technology}          & CMOS                  \\ \cline{1-2}
	\textbf{Image size/resolution}      & 2048 x 2048 pixels    \\ \cline{1-2}
	\textbf{Maximum frame rate}         & 89 FPS                \\ \cline{1-2}
	\textbf{Operating temperature}      & 0 to 50 °C            \\ \cline{1-2}
	\textbf{Storage temperature}        & -30 to 65 °C          \\ \cline{1-2}
	\textbf{Operating/storage humidity} & 0 to 85\%             \\ \cline{1-2}
	\textbf{Filter peak wavelength}     & 940 nm                \\ \cline{1-2}
	\textbf{Lens focal length}          & 16 mm                 \\ \cline{1-2}
	\end{tabular}
\end{table}

\textbf{Fingerprint scanners:} Two fingerprint scanners were used to capture images of the data subjects' fingerprints: (a) the thermal OYSTER III scanner from NEXT Biometrics; and (b) the optical CSD101i scanner from Thales. Fig. 3 presents images of these scanners, and Table II summarizes the scanners' technical specifications. 

\begin{figure}[!h]
	\centering
	\includegraphics[width=0.9\columnwidth]{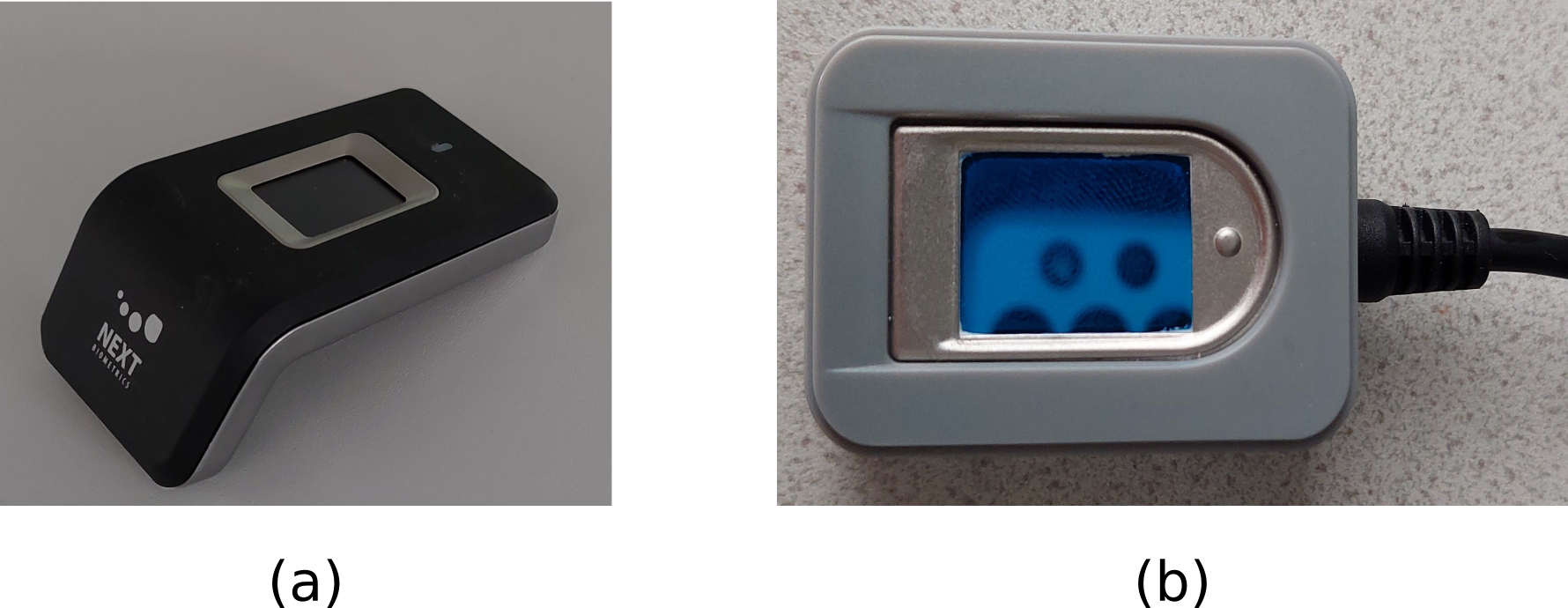}
	\caption{Fingerprint scanners: (a) OYSTER III; (b) CSD101i.}
\end{figure}

\begin{table}[!h]
	\centering
	\caption{Fingerprint scanner specifications.}
	\begin{tabular}{|l|c|c|}
	\hline
	\textbf{Manufacturer}               & NEXT Biometrics   & Thales           \\ \hline
	\textbf{Model}                      & OYSTER III        & CSD101i          \\ \hline
	\textbf{Dimensions}                 & 85 x 40 x 28.5 mm & 46 x 32 x 15 mm  \\ \hline
	\textbf{Active sensing area}        & 15.24 x 20.32 mm  & 18 x 12.8 mm     \\ \hline
	\textbf{Sensor technology}          & Thermal           & Optical          \\ \hline
	\textbf{Image resolution}           & 508 ppi           & 500 ppi          \\ \hline
	\textbf{Image size}                 & 300 x 400 pixels  & 256 x 360 pixels \\ \hline
	\textbf{Image capture time}         & 0.8 sec           & \textless 2 sec  \\ \hline
	\textbf{Operating temperature}      & -10 to 50 °C      & 0 to 55 °C       \\ \hline
	\textbf{Storage temperature}        & -20 to 70 °C      & -20 to 55 °C     \\ \hline
	\textbf{Operating/storage humidity} & 95\%              & 10-90\%          \\ \hline
	\textbf{Sensor hardness}            & \textgreater 9H   & 9H               \\ \hline
	\end{tabular}
\end{table}

\textbf{Microphones:} Two microphones were used to capture audio samples of the data subjects' voices: (a) the ME72 model from Valeo; and (b) the AT2020 model from Audio-Technica. Fig. 4 shows images of these microphones, and Table III summarizes the microphones' technical specifications. 

\begin{figure}[!h]
	\centering
	\includegraphics[width=0.9\columnwidth]{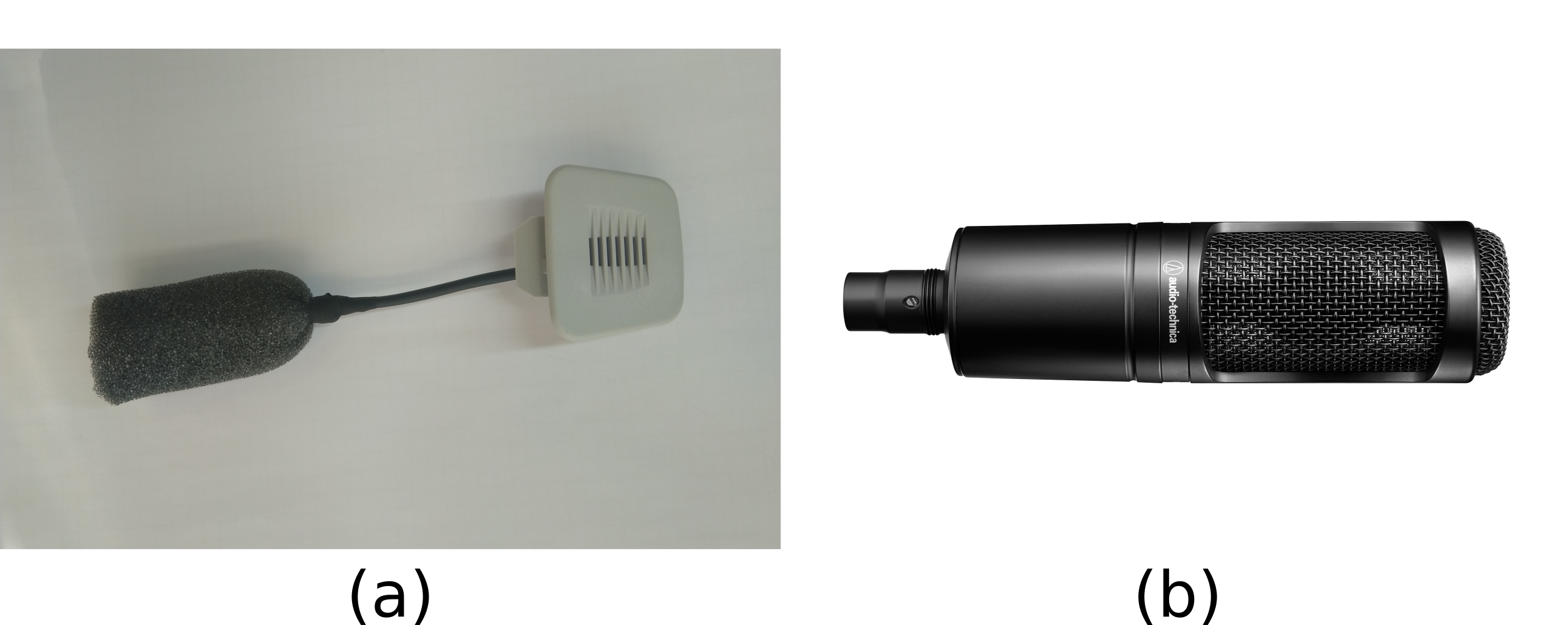}
	\caption{Microphones: (a) ME72; (b) AT2020.}
\end{figure}

\begin{table}[!h]
	\centering
	\caption{Microphone specifications.}
	\begin{tabular}{|l|c|c|}
	\hline
	\textbf{Manufacturer}             & Valeo & Audio-Technica \\ \hline
	\textbf{Model}                    & ME72  & AT2020         \\ \hline
	\textbf{Signal to Noise (dB(A))}  & \textgreater 65  & 74             \\ \hline
	\textbf{Current consumption (mA)} & 6 ± 3 & 2              \\ \hline
	\end{tabular}
\end{table}

While the AT2020 microphone worked as is, the ME72 had to be connected to a special circuit to get the audio signal from the microphone into the laptop used for data collection. Fig. 5 shows the circuit schematic, and Fig. 6 shows a real-life image of the ME72 microphone connected to this circuit. 

\begin{figure}[!h]
	\centering
	\includegraphics[width=0.9\columnwidth]{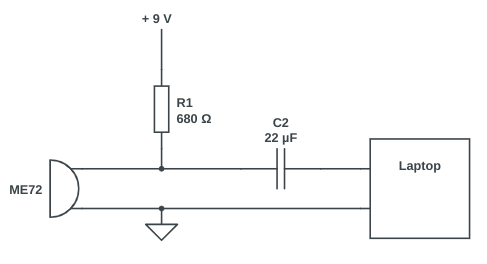}
	\caption{Schematic of the circuit connecting the ME72 microphone to a battery and a laptop.}
\end{figure}

\begin{figure}[!h]
	\centering
	\includegraphics[width=0.9\columnwidth]{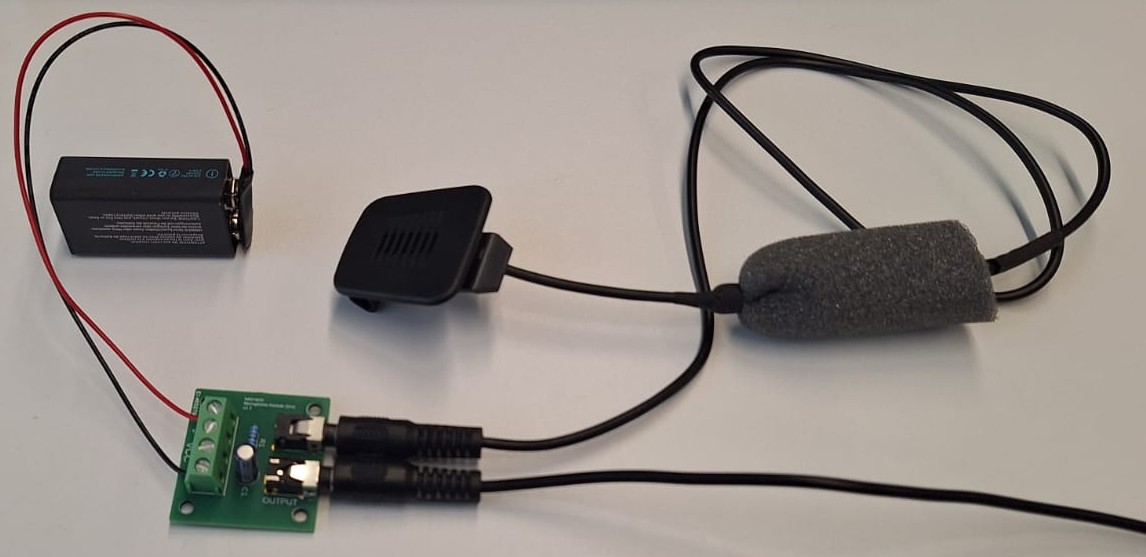}
	\caption{ ME72 microphone, with its circuit and power supply. The circuit's second output connects to the laptop's audio jack.}
\end{figure}

\subsection{Data collection protocols}

The first step of the data collection was to gather a set of 200 volunteers for each piece of biometric data (i.e., face, fingerprint, and voice). The call for volunteers was propagated by sending e-mails to various mailing lists and by word of mouth. Every effort was made to attract and select a diverse set of data subjects, in order to achieve as close a balance as possible between the genders, skin colours, and ages of the participants, thereby reducing (as much as possible) demographic bias in the acquired data. Furthermore, before collecting any data, each data subject was asked to sign a consent form. The data subjects were also informed that they have the right to withdraw their data at any time. Finally, at the end of the data collection, the participants were renumerated for their time and effort. So, the data collection process was fully consensual, ethical, and legal, and the acquired data is safeguarded by Swiss laws and regulations on data protection.  

Once the volunteers were gathered and the consent forms signed, the next step was to collect the volunteers' biometric data (face, fingerprint, and voice). The data collection was performed inside the test car, using the sensors described in Section II.A. The data was captured during two different sessions: one while the car was parked indoors (in a garage), and the other while the car was parked outdoors. Different variabilities were deliberately incorporated into the data collection to simulate ``noise'' (i.e., non-ideal or ``unclean'' biometric data) that may be encountered in practice during driver identification. The data collection protocol for each of the three considered biometric modalities is described below. 

\begin{center}
\underline{Face} 
\end{center}

Each volunteer was randomly assigned an integer ID between 1 and 200, and this ID was associated with the acquired face data (the data subjects' names were not recorded). Apart from this unique ID, the following additional (demographic and environmental) information was recorded for each data subject and each capture session: 

\begin{itemize}

\item Gender (binary): \texttt{\{Male, Female\}}
\item Skin colour (Fitzpatrick scale): \texttt{\{Types I-VI\}}
\item Age (category):
 
	\begin{itemize}
	\item \texttt{1: 18-24 years old}
	\item \texttt{2: 25-29 years old}
	\item \texttt{3: 30-34 years old}
	\item \texttt{4: 35-39 years old}
	\item \texttt{5: 40-49 years old}
	\item \texttt{6: 50-59 years old}
	\item \texttt{7: 60+ years old}
	\end{itemize}

\item Facial hair: \texttt{\{None, Moustache, Goatee, Beard\}}
\item Accessories: \texttt{\{None, Glasses, Sunglasses, Mask, Hat\}}
\item Weather: \texttt{\{Rainy, Cloudy, Sunny\}} 

\end{itemize}

Then, the data subject was asked to sit in the driver's seat of the car, and videos of their face were acquired (using the camera) in the following scenarios, to incorporate different variabilities into the recorded face data: 

\begin{enumerate}

\item \textbf{Indoors:} The car was parked inside a garage, with controlled (artificial) lighting.

	\begin{enumerate}[(i)]
	
	\item Subject wore \textit{no} deliberate accessories. 
	\item Subject wore a \textit{hygienic mask}. 
	\item Subject wore a \textit{hat} (cap). 
	
	\end{enumerate}	 
	
\item \textbf{Outdoors:} The car was parked outside, with uncontrolled (natural) lighting.  

	\begin{enumerate}[(i)]
	
	\item Subject wore \textit{no} deliberate accessories.
	\item Subject wore \textit{sunglasses}. 
	\item Subject wore a \textit{hat} (cap). 
	
	\end{enumerate}

\end{enumerate}

Fig. 7 and Fig. 8 show examples of face images (frames extracted from videos) obtained from the \textbf{Indoors} and \textbf{Outdoors} scenarios, with the different variations incorporated. 

\begin{figure}[!h]
	\centering
	\includegraphics[width=\columnwidth]{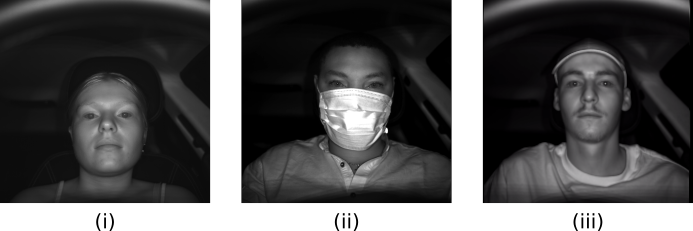}
	\caption{Examples of face images in the Indoors scenario: (i) no accessories; (ii) hygienic mask; (iii) hat.}
\end{figure}

\begin{figure}[!h]
	\centering
	\includegraphics[width=\columnwidth]{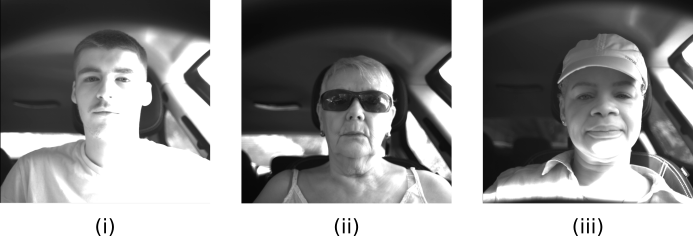}
	\caption{Examples of face images in the Outdoors scenario: (i) no accessories; (ii) sunglasses; (iii) hat.}
\end{figure}

The required accessories were provided to the subjects. When possible, subjects also brought their own sunglasses or hats, adding variation to the data with these accessories. 

For each of the acquisition scenarios described above, face videos of the data subject were captured while the subject: 

\begin{enumerate}[(a)]

\item Remained \textit{still}, with a \textit{neutral} facial expression (5 sec).
\item Remained \textit{still}, with a \textit{neutral} facial expression, and with \textit{eyes closed} (5 sec). 
\item Performed natural \textit{face/head movements}, e.g., turning, talking, smiling, laughing (15 sec). 

\end{enumerate}

Fig. 9 shows examples of face images where the subject is performing these actions.

\begin{figure}[!h]
	\centering
	\includegraphics[width=\columnwidth]{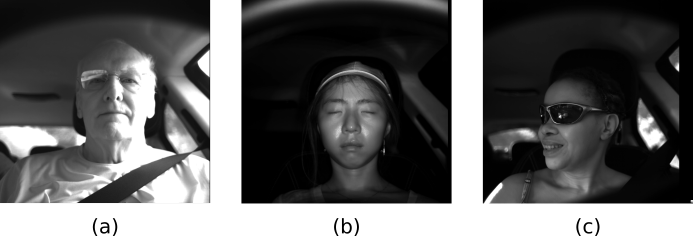}
	\caption{Examples of face images where the data subject: (a) is still with a neutral facial expression; (b) has eyes closed; (c) is performing natural face/head movements.}
\end{figure}

So, in total, the \textit{iCarB-Face} dataset consists of 18 face videos per data subject (i.e., 3 videos for each of the 6 described scenarios: indoors with no accessories + indoors while wearing a hygienic mask + indoors while wearing a hat + outdoors with no accessories + outdoors while wearing sunglasses + outdoors while wearing a hat). This amounts to a total of \textbf{3,600 face videos} across all 200 data subjects. 

\begin{center}
\underline{Fingerprints}
\end{center}

Each volunteer was randomly assigned an integer ID between 1 and 200, and this ID was associated with the acquired fingerprint data (the data subjects' names were not recorded). Apart from this unique ID, the following additional (environmental) information was recorded for each data subject and each capture session: 

\begin{itemize}

\item Temperature inside the vehicle: \texttt{Float (in $^{\circ}$C)} 

\end{itemize}

Then, the data subject was asked to sit in the driver's seat of the car, select one finger and present it to each of the two fingerprint scanners in turn to capture the fingerprint images.  The fingerprint images were acquired in two scenarios: 

\begin{enumerate}

\item \textbf{Indoors:} The car was parked inside a garage, with controlled (artificial) lighting.
\item \textbf{Outdoors:} The car was parked outside, with uncontrolled (natural) lighting.  

\end{enumerate}

In both scenarios, the following types of fingerprint images were captured to incorporate different variabilities into the recorded fingerprint data, using each of the two fingerprint scanners in turn: 

\begin{enumerate}[(i)]

\item \textit{Normal:} Fingerprint was captured as is (4 images). 
\item \textit{Dry:} Finger was dried using hand disinfectant and a paper towel (2 images).
\item \textit{Moist:} Finger was moistened by rubbing it on the nose or forehead (2 images). 
\item \textit{Hot:} Finger was heated by placing it on a hot-water bottle (2 images).
\item \textit{Cold:} Finger was cooled by placing it on an ice block (2 images). 

\end{enumerate}

Fig. 10 and Fig. 11 show examples of fingerprint images with the different variabilities described above, captured using the OYSTER III and CSD101i fingerprint scanners.

\begin{figure}[!h]
	\centering
	\includegraphics[width=\columnwidth]{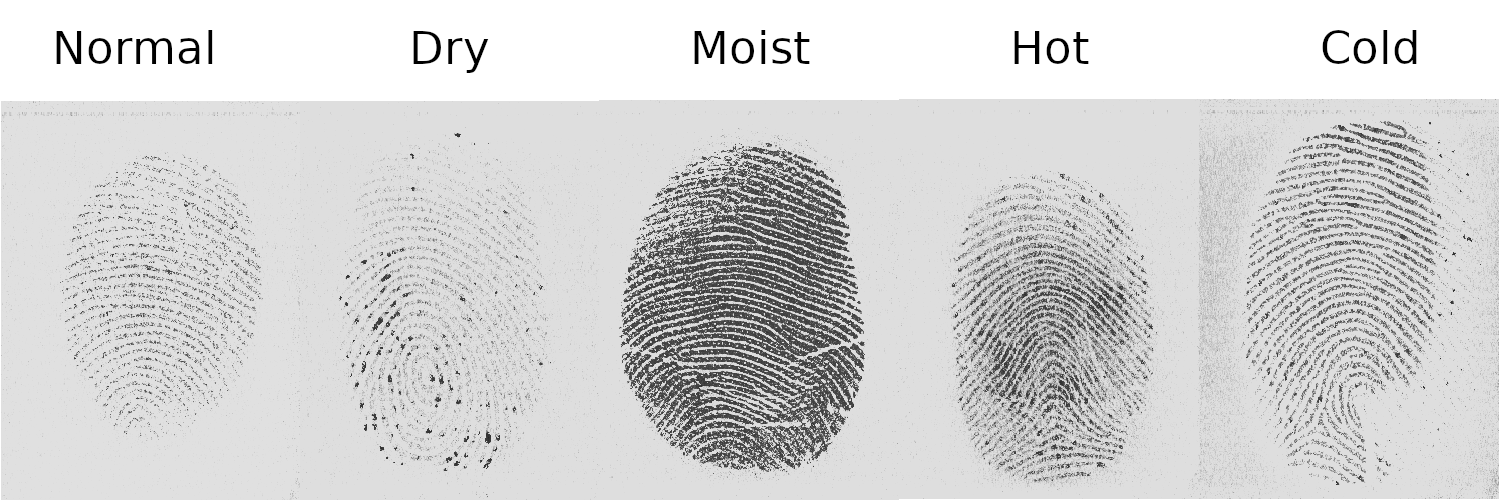}
	\caption{Examples of fingerprint images acquired with the Oyster III scanner.}
\end{figure}

\begin{figure}[!h]
	\centering
	\includegraphics[width=\columnwidth]{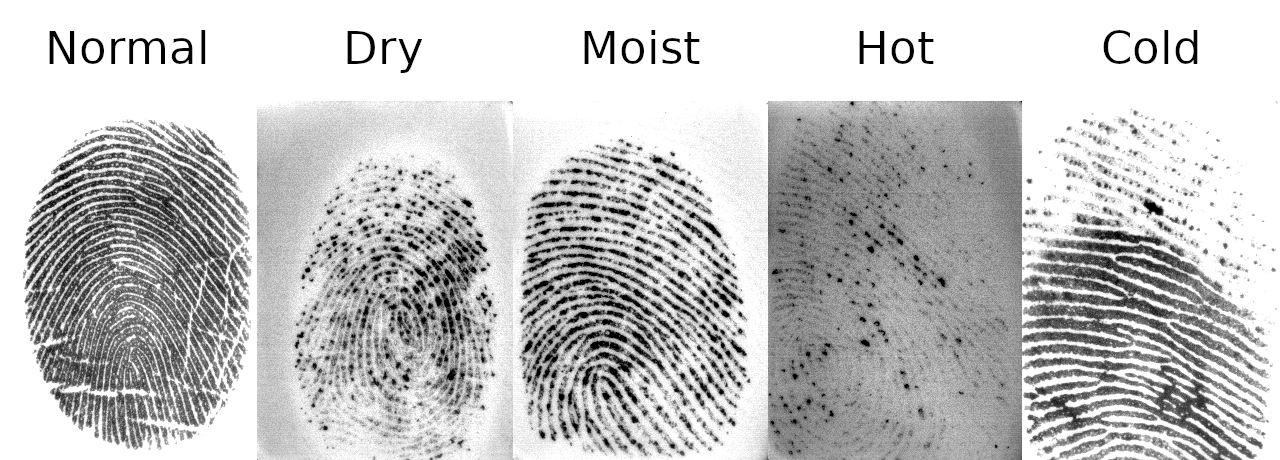}
	\caption{Examples of fingerprint images acquired with the CSD101i scanner. }
\end{figure}

Note that the fingerprint scanners were cleaned between the different sets of fingerprint captures (i.e., at the start, then after each set of the ``normal'', ``dry'', ``moist'', ``hot'', and ``cold'' captures). Furthermore, the ``dry'', ``moist'', ``hot'', and ``cold'' simulations were renewed when switching between the two scanners (e.g., the finger was re-heated). 

So, in total, the \textit{iCarB-Fingerprint} dataset consists of 48 fingerprint images per data subject (i.e., 4 normal + 2 dry + 2 moist + 2 hot + 2 cold (so 12 in total), separately for each of the 2 fingerprint scanners and each of the 2 capture sessions (indoors and outdoors)), for most data subjects. For a few subjects, it was not possible to acquire fingerprint images in some scenarios, so in those cases the obtained number of fingerprint images was less than 48. Taking this point into consideration, the dataset consists of a total of \textbf{9,521 fingerprint images} across all 200 data subjects. 

\begin{center}
\underline{Voice}
\end{center}

Each volunteer was randomly assigned an integer ID between 1 and 200, and this ID was associated with the acquired voice data (the data subjects' names were not recorded). Apart from this unique ID, the following additional (demographic and environmental) information was recorded for each data subject and each capture session: 

\begin{itemize}

\item Gender (binary): \texttt{\{Male, Female\}} 
\item Language: \texttt{\{English, Other\}}

\end{itemize}

Then, the data subject was asked to sit in the driver's seat of the vehicle, such that samples of their voice could be acquired using the two microphones (simultaneously).  The voice samples were acquired in the following scenarios, the aim of which was to incorporate different variabilities into the recorded voice data:     

\begin{enumerate}

\item \textbf{Indoors:} The car was parked inside a garage, and various background noises were simulated. The data subjects were asked to read a series of 30 sentences in English (or in another language if they were unable to speak English).   

	\begin{enumerate}[(i)]
	
	\item \textit{Noiseless:} Car doors and windows were closed, and there were no deliberate background noises (30 sentences). 
	
	\item \textit{External noises:} Traffic noises (e.g., car horns, sounds of passing cars) were played on a smartphone (placed outside the car) and emanated through speakers placed near the driver's door, while the driver's window was closed (5 sentences) then open (5 sentences). 
	
	\item \textit{Internal noises:} Different noises were played on a smartphone inside the car, including classical music (5 sentences), rock music (5 sentences), background discussion (5 sentences), and impulsive noises like dogs barking and babies crying (5 sentences). 
	
	\end{enumerate}

\item \textbf{Outdoors:} The car was parked outside, and there was no noise simulation (i.e., any background noises were natural). The data subjects were asked to read or make up a series of 30 sentences in a language other than English (unless English was their only language). 

	\begin{enumerate}[(i)]
	
	\item \textit{Window closed:} The driver's window was closed (15 sentences). 
	\item \textit{Window open:} The driver's window was open (15 sentences). 
	
	\end{enumerate}

\end{enumerate}

So, in total, the \textit{iCarB-Voice} dataset consists of 180 voice samples per data subject. This amounts to a total of \textbf{36,000 voice samples} across all 200 data subjects.

\section{Data Description}

The three datasets presented in this paper, \textit{iCarB-Face}, \textit{iCarB-Fingerprint}, and \textit{iCarB-Voice}, contain biometric data from 200 human volunteers. The selection of these data subjects aimed for gender balance, resulting in equal representation of male and female participants (i.e., 100 males and 100 females). Additionally, the datasets encompass a diverse array of skin tones and age groups, as shown in Fig. 12. 

\begin{figure}[!h]
	\centering
	\includegraphics[width=\columnwidth]{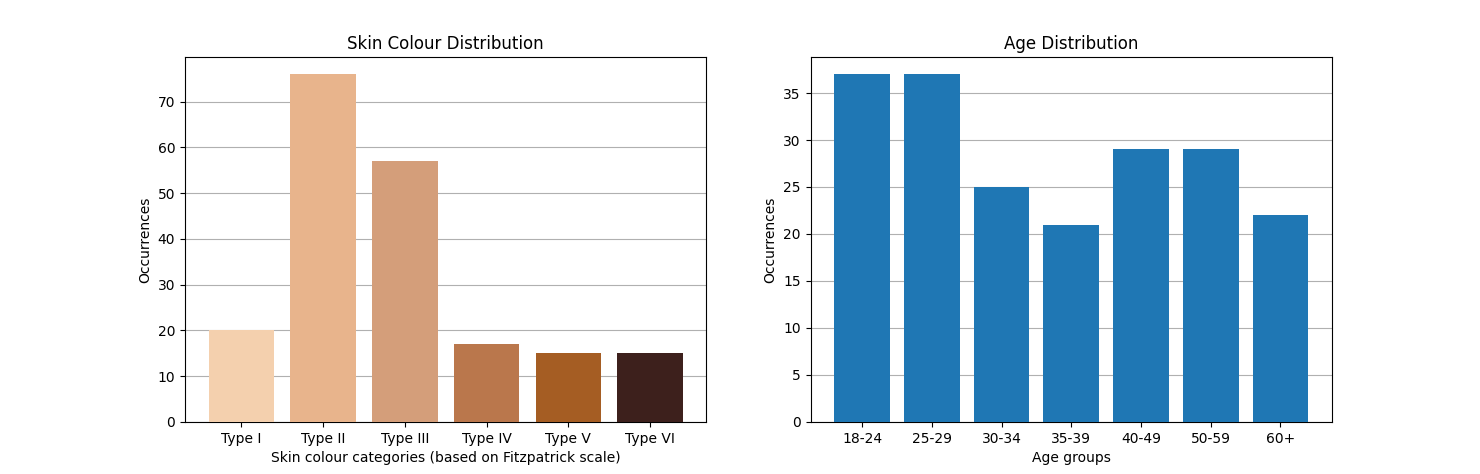}
	\includegraphics[width=\columnwidth]{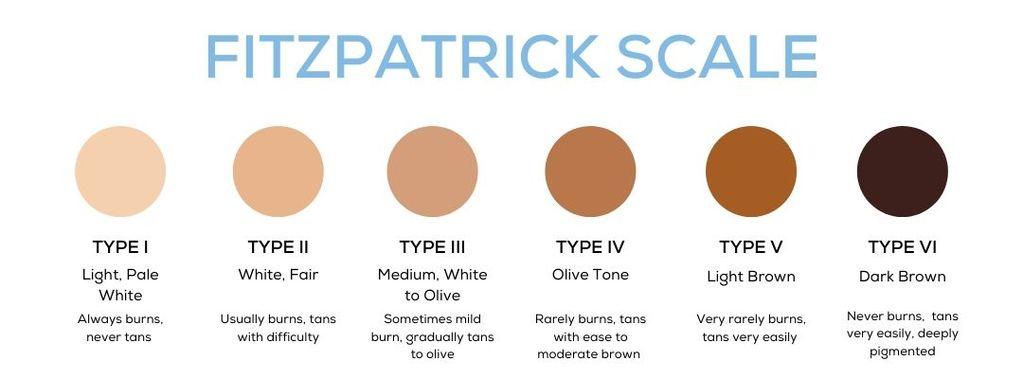}
	\caption{Dataset statistics, with the skin colour distribution (based on the Fitzpatrick scale) on the left and the categorized age distribution on the right.}
\end{figure}

The number of files per dataset depends on the data collection protocol (e.g., selected variations) for each modality, as described in Section II. Table IV gives an overview of the three datasets with the amount of data they contain. 

\begin{table*}[!h]
	\centering
	\caption{Overview of the three datasets with the amount of data they contain.}
	\begin{tabular}{|l|c|c|c|c|c|}
	\hline
	\multicolumn{1}{|c|}{\textbf{Dataset name}} &
	  \textbf{File type} &
	  \textbf{\# of subjects} &
	  \textbf{\# of files (expected)} &
	  \textbf{\# of files (obtained)} &
	  \textbf{Size (GB)} \\ \hline
	iCarB-Face        & .avi & 200 & 3600  & 3600  & 92.1 \\ \hline
	iCarB-Fingerprint & .bmp & 200 & 9600  & 9521 & 0.97 \\ \hline
	iCarB-Voice       & .wav & 200 & 36000 & 36000 & 28.5 \\ \hline
	\end{tabular}
\end{table*}

The three datasets share a similar structure and provide the following information alongside the biometric data: 

\begin{itemize}

\item Multiple pre-defined evaluation protocols. These protocols can be used to evaluate a given biometric system on the dataset, either globally or by focusing on a specific variation (e.g., dry fingerprints, noisy speech). When applicable, a protocol will use data coming from a single sensor; no intra-sensor variations were considered. 

\item Metadata describing the different subjects (e.g., facial hair, gender) or the different acquisition scenarios (e.g., weather, temperature). As described in Section II, the provided metadata varies depending on the dataset. 

\item General information about the dataset, describing: (i) the data it contains, (ii) the different variations, (iii) how filenames were constructed, (iv) the provided metadata, and (v) the different evaluation protocols. In this document we also report any issues with the data. 

\end{itemize}

Fig. 13 illustrates the directory structure for each dataset. 

\begin{figure}[!h]
	\centering
	\includegraphics[width=0.9\columnwidth]{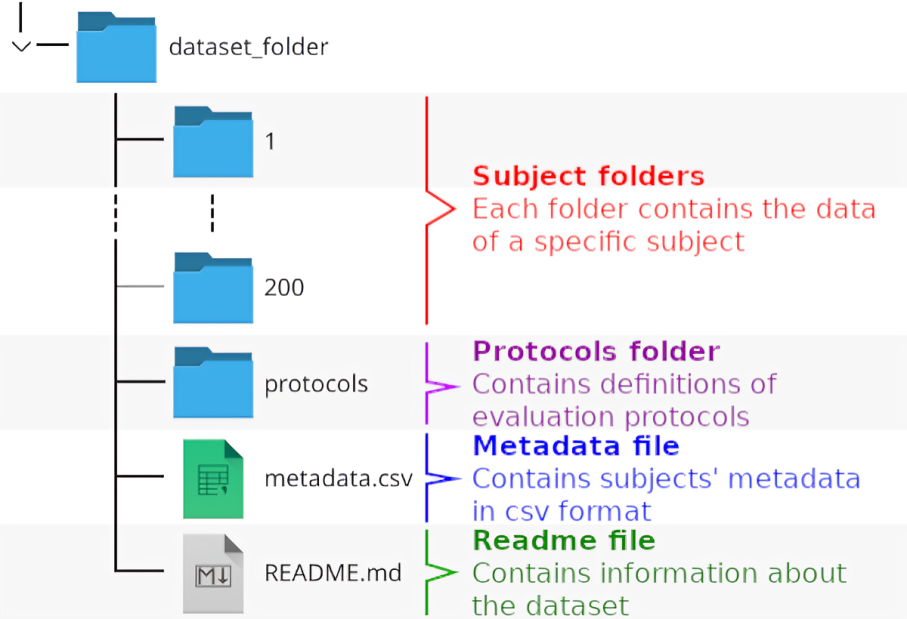}
	\caption{Individual dataset file structure.}
\end{figure}

Biometric data for each subject are stored in a dedicated folder labelled with the subject's unique ID, referred to as \textbf{\texttt{\textcolor{red}{<usr\_id>}}}, which ranges from 1 to 200. All data from both the indoor and outdoor sessions (see Section II for details on the data collection protocols) are contained within this folder. Files from different sessions are distinguished by the \textbf{\texttt{\textcolor{orange}{<ses\_id>}}} tag included in the filename. The filename structure changes based on the dataset to indicate (i) the sensor used, and (ii) the variation performed, which can be identified directly from the filename. The sub-sections below detail the filename structure for each dataset. 

\subsection{iCarB-Face}

This dataset contains face videos of data subjects sitting in the driver’s seat of the car. The dataset's filename structure is:  

\medskip
\begin{center}
\noindent \textbf{\texttt{./\textcolor{red}{<usr\_id>}/\textcolor{red}{<usr\_id>}\_\textcolor{orange}{<ses\_id>}\_\\\textcolor{blue}{<acc\_id>}\_\textcolor{magenta}{<act\_id>}.avi}}
\end{center}
\medskip

Where: 

\begin{itemize}

\item \texttt{\textbf{\textcolor{blue}{<acc\_id>}}} refers to the kind of accessory used for the capture (0 for no accessories, 1 for a hygienic mask, 2 for a hat, 3 for sunglasses). 

\item \texttt{\textbf{\textcolor{magenta}{<act\_id>}}} indicates the action performed by the subject during the recording (0 for frontal face view, 1 for eyes closed, 2 for natural face/head movement). 

\end{itemize}

For example, file \textbf{\texttt{./\textcolor{red}{50}/\textcolor{red}{50}\_\textcolor{orange}{1}\_\textcolor{blue}{1}\_\textcolor{magenta}{2}.avi}} contains a video recording of \textcolor{red}{subject 50} during \textcolor{orange}{session 1}. During this recording, the subject was wearing a \textcolor{blue}{hygienic mask} and performing \textcolor{magenta}{natural face/head movements} in front of the camera. 

\subsection{iCarB-Fingerprint}

This dataset contains fingerprint images acquired using two different fingerprint scanners. The filenames are structured as follows: 

\medskip
\begin{center}
\noindent \textbf{\texttt{./\textcolor{red}{<usr\_id>}/\textcolor{red}{<usr\_id>}\_\textcolor{orange}{<ses\_id>}\_\\\textcolor{blue}{<scr\_id>}\_\textcolor{magenta}{<var\_id>}\_\textcolor{cyan}{<cap\_id>}.bmp}}
\end{center}
\medskip

Where: 

\begin{itemize}

\item \textbf{\textcolor{blue}{\texttt{<scr\_id>}}} is the ID of the fingerprint scanner used for the capture (0 for OYSTER III from NEXT Biometrics, 1 for CSD101i from Thales). 

\item \textbf{\textcolor{magenta}{\texttt{<var\_id>}}} is the variation performed on the captured fingerprint (0 for normal, 1 for dry, 2 for moist, 3 for hot, 4 for cold). 

\item \textbf{\textcolor{cyan}{\texttt{<cap\_id>}}} is the capture index (an integer between 0 and 3). 

\end{itemize}

For example, the \textbf{\texttt{./\textcolor{red}{127}/\textcolor{red}{127}\_\textcolor{orange}{2}\_\textcolor{blue}{0}\_\textcolor{magenta}{3}\_\textcolor{cyan}{1}.bmp}} file contains the \textcolor{cyan}{second fingerprint image} acquired with the \textcolor{blue}{OYSTER III scanner} of \textcolor{red}{subject 127} during their \textcolor{orange}{second session}. The filename also indicates that the subject’s finger was \textcolor{magenta}{hot}. 

\subsection{iCarB-Voice}

This dataset contains voice recordings acquired with two different microphones. The filename structure is as follows: 

\medskip
\begin{center}
\noindent \textbf{\texttt{./\textcolor{red}{<usr\_id>}/\textcolor{red}{<usr\_id>}\_\textcolor{orange}{<ses\_id>}\_\\\textcolor{blue}{<mic\_id>}\_\textcolor{magenta}{<win\_id>}\_\textcolor{cyan}{<noi\_id>}\_\textcolor{purple}{<snt\_id>}.wav}}
\end{center} 
\medskip

Where: 

\begin{itemize}

\item \textbf{\textcolor{blue}{\texttt{<mic\_id>}}} represents the ID of the microphone used for the capture (0 for AT2020 from Audio-Technica, 1 for ME72 from Valeo). 

\item \textbf{\textcolor{magenta}{\texttt{<win\_id>}}} indicates whether the car’s windows were closed (0) or open (1). 

\item \textbf{\textcolor{cyan}{\texttt{<noi\_id>}}} is the ID of the external noise added to the capture: 

	\begin{itemize}
	
	\item 0 for no artificial noise. 

    \item 1 for traffic noise. 

    \item 2 for classical music. 

    \item 3 for rock music. 

    \item 4 for background discussion. 

    \item 5 for impulsive noise. 	
	
	\end{itemize}
	
\item \textbf{\textcolor{olive}{\texttt{<snt\_id>}}} is the spoken sentence index (an integer between 0 and 29). 

\end{itemize}

For example, the \textbf{\texttt{./\textcolor{red}{193}/\textcolor{red}{193}\_\textcolor{orange}{1}\_\textcolor{blue}{1}\_\textcolor{magenta}{1}\_\textcolor{cyan}{3}\_\textcolor{olive}{25}.wav}} file contains a voice recording of \textcolor{red}{subject 193} during their \textcolor{orange}{first session}. The \textcolor{blue}{ME72 microphone} was used, the \textcolor{magenta}{car’s windows were open}, and \textcolor{cyan}{rock music }was played in the background. The filename also indicates that, during this recording, the subject was reading sentence number \textcolor{olive}{25}.  

\subsection{Using the provided evaluation protocols}

As previously mentioned, each dataset includes a protocols folder that contains a set of pre-defined evaluation protocols. These protocols are detailed in two CSV files: \texttt{for\_enrolling.csv} describes the biometric samples used to create a reference database, and \texttt{for\_probing.csv} lists the samples used to probe (test) this database. Each row in these CSV files provides details about a specific sample, including (i) its path relative to the dataset's root, (ii) the subject identifier, (iii) the template identifier, and (iv) additional metadata. The subject identifier indicates the owner of the biometric sample, while the template identifier tells the biometric system where to store the extracted biometric features. If two biometric samples share the same template identifier, the system will combine the features from these samples into a single model. Fig. 14 describes the structure of the protocols folder. 

\begin{figure}[!h]
	\centering
	\includegraphics[width=\columnwidth]{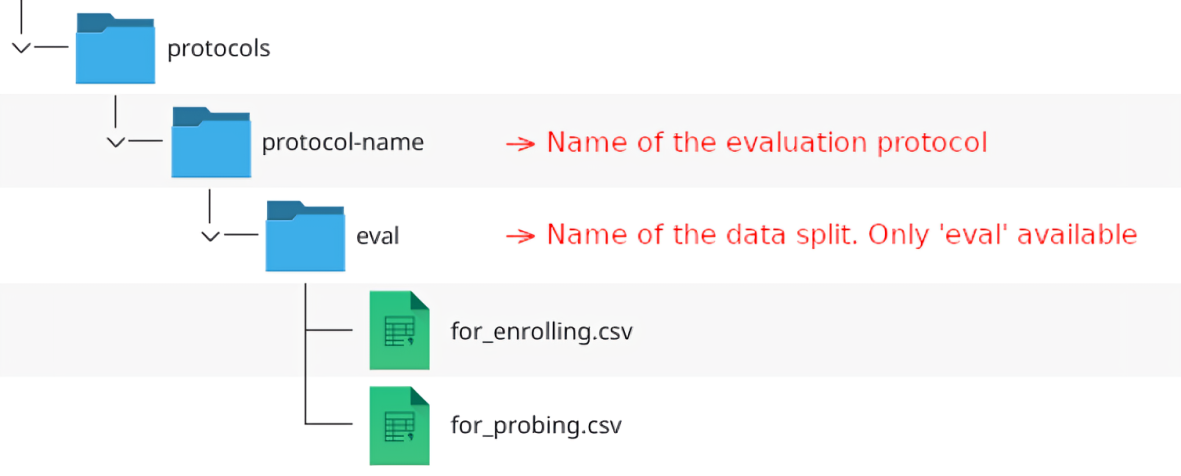}
	\caption{Protocols folder structure.}
\end{figure}

The protocol name indicates which variation is considered in the protocol; for example, in \textit{iCarB-Face}, \textbf{indoor-all} will evaluate all the biometric samples that were acquired indoors, while \textbf{indoor-mask-frontal} will evaluate only the indoor samples where the subject was wearing a hygienic mask and looking straight at the camera.

\section{Limitations}

This section describes the limitations of our three datasets: \textit{iCarB-Face}, \textit{iCarB-Fingerprint}, and \textit{iCarB-Voice}.

\subsubsection{iCarB-Face} 

The second data acquisition session occurred outdoors with uncontrolled lighting, which resulted in some face captures being overexposed or underexposed. This variation was intentional (i.e., we deliberately chose not to control the lighting for the outdoor session), to simulate face acquisition challenges in a real-life driver recognition scenario.  However, the overexposed and underexposed images may present a significant challenge for some face recognition systems (e.g., problems with face detection). 

\subsubsection{iCarB-Fingerprint} 

We experienced challenges in collecting fingerprint images from certain individuals, particularly older subjects and subjects with dry skin. Both fingerprint scanners were affected by these issues, and any missing fingerprint images have been documented in the README file accompanying the dataset. The optical sensor (Thales CSD101i) also appeared to be sensitive to ambient brightness, yielding lower quality images under strong lighting conditions. 

\subsubsection{iCarB-Voice} 

During part of the second data acquisition session, which took place outdoors, the driver's car window was open. On certain days, it was quite windy outside, which caused the speech signal to become distorted and thus challenging to identify. Again, this variation was intentional, so no specific actions were taken to control for the noise.  Consequently, some voice recognition systems may struggle with the noisy data.

\section{Ethics Statement}
  
All participants in this data collection were volunteers, and each participant signed a consent form to allow for their data to be collected and used for the specified research purposes. The project under which this data was collected, was approved by Idiap’s DREC under internal reference number 2022-06-14/78. 

\section*{Acknowledgment}

We would like to express our gratitude to all the volunteers who participated in this data collection effort. Furthermore, we would like to acknowledge our industrial partner (whose name has been omitted as per their request due to NDA restrictions), as well as the Swiss Center for Biometrics Research and Testing, for funding the collection of the three datasets presented in this paper (\textit{iCarB-Face}, \textit{iCarB-Fingerprint}, and \textit{iCarB-Voice}). 

\bibliographystyle{IEEEtran}
\bibliography{bibliography}

\end{document}